\documentclass[a4paper,fleqn]{cas-dc}

\usepackage{soul} 
\usepackage{color, xcolor} 
\usepackage[numbers]{natbib}
\usepackage[boxed,commentsnumbered,ruled,linesnumbered]{algorithm2e}
\usepackage{tabularx}
\usepackage{bbding}
\usepackage{makecell}

\usepackage{hyperref}
\hypersetup{
    colorlinks=true,
    linkcolor=blue,
    urlcolor=blue,
    anchorcolor=blue,
    citecolor=blue
}

\usepackage[switch]{lineno}

\usepackage{soul}
\usepackage{color, xcolor}  
\sethlcolor{yellow}
\soulregister{\cite}7 
\soulregister{\citep}7 
\soulregister{\citet}7 
\soulregister{\ref}7 
\soulregister{\pageref}7 
\soulregister{\underline}7 

\begin{document}
\shorttitle{Large Language Models for Medicine: A Survey} 
\shortauthors{Y. Zheng \textit{et al.}}


\title [mode = title]{Large Language Models for Medicine: A Survey}   
\author[1]{Yanxin Zheng}
\ead{DuoDuOoozyx@gmail.com}
\address[1]{College of Cyber Security, Jinan University, Guangzhou 510632, China}

\author[1]{Wensheng Gan}
\cortext[cor1]{Corresponding author}
\ead{wsgan001@gmail.com}
\cormark[1]

\author[1]{Zefeng Chen}
\ead{czf1027@gmail.com}

\author[2]{Zhenlian Qi}
\ead{qzlhit@gmail.com}
\address[2]{School of Information Engineering, Guangdong Eco-Engineering Polytechnic, Guangzhou 510520, China}

\author[3]{Qian Liang}
\ead{Liangqian0914@163.com}
\address[3]{Shenzhen People's Hospital (The Second Clinical Medical College, Jinan University), Shenzhen 518020, China}

\author[4]{Philip S. Yu}
\ead{psyu@uic.edu}
\address[4]{Department of Computer Science, University of Illinois Chicago, Chicago, IL 60607, USA}

\begin{abstract}
   To address challenges in the digital economy's landscape of digital intelligence, large language models (LLMs) have been developed. Improvements in computational power and available resources have significantly advanced LLMs, allowing their integration into diverse domains for human life. Medical LLMs are essential application tools with potential across various medical scenarios. In this paper, we review LLM developments, focusing on the requirements and applications of medical LLMs. We provide a concise overview of existing models, aiming to explore advanced research directions and benefit researchers for future medical applications. We emphasize the advantages of medical LLMs in applications, as well as the challenges encountered during their development. Finally, we suggest directions for technical integration to mitigate challenges and potential research directions for the future of medical LLMs, aiming to meet the demands of the medical field better.
\end{abstract}


\makeatletter\def\Hy@Warning#1{}\makeatother
\maketitle

\section{Introduction}  \label{sec: introduction}

The medical domain experiences a myriad of challenges, driven by the rapid growth of data and the need for improved patient care and medical research \cite{mccue2017scope}. As traditional methods struggle to cope with the vast volumes of information and intricate medical terminology, artificial intelligence (AI) technologies have emerged as pivotal tools in addressing these complexities \cite{nilsson1982principles}. AI methodologies have played a significant role in revolutionizing medical information retrieval and processing. Information needs in medicine refer to the relevant information required by medical professionals, patients, and researchers in areas such as clinical practice, medical research, and health management \cite{silberg1997assessing, duggan2008medicine}. This includes case data, medical knowledge, treatment guidelines, drug information, and the latest research findings in disease prevention and health promotion. Information plays a crucial role in the medical field, contributing to accurate diagnoses, effective treatment of diseases, improved patient care, and advancements in medical research \cite{waitzkin1985information}. Nevertheless, despite significant advancements, existing medical information retrieval and processing systems encounter formidable obstacles. The sheer magnitude and dynamic nature of medical data present challenges for conventional search engines and databases, impairing their ability to deliver swift and accurate results. Moreover, the intricate domain-specific language and evolving medical knowledge further exacerbate these limitations, often resulting in inaccuracies and inefficiencies in information retrieval \cite{carpineto2012survey}.

In recent years, with the development of web 3.0 \cite{gan2023web,wan2023web3} and Internet of behaviors \cite{sun2023internet}, the emergence of large language models (LLMs) \cite{gan2023large, shanahan2024talking} has represented a significant breakthrough in the fields of artificial intelligence \cite{xi2023rise} and data sciences \cite{zhang2022tusq,gan2022discovering}. LLMs, based on deep learning, are specifically designed for processing and generating natural language text. They acquire language patterns and knowledge through pre-training on massive text datasets, resulting in strong performance across various natural language processing (NLP) tasks. The development of LLMs can be attributed to two crucial factors \cite{zhao2023survey}. Firstly, the availability of large-scale pre-training datasets enables LLMs to acquire extensive language knowledge and patterns by leveraging a vast amount of text data from the Internet. Secondly, advancements in computational resources and technology have provided the necessary foundation for training and inference with large-scale language models. Extensive training empowers LLMs with robust reasoning capabilities, enabling them to generalize effectively and adapt to text data from diverse domains and tasks. Consequently, LLMs can infer and generate coherent and natural text, effectively handling complex semantics and language structures. By extensively learning language knowledge during the pre-training phase, LLMs retain certain capabilities to tackle unseen tasks or domains. Their zero-shot learning ability equips them with a level of versatility and adaptability. To some extent, LLMs fulfill conversational functionalities. One notable example is the generative pre-trained Transformer (GPT) model series, which adopts the Transformer architecture \cite{lin2022survey}. Transformer is a deep learning model based on attention mechanisms and demonstrates exceptional proficiency in NLP tasks. Based on the aforementioned factors, LLMs typically exhibit characteristics such as large-scale pre-training, strong generalization, contextual understanding, text generation, zero-shot learning, and certain interactive capabilities. Due to technological advancements and increased accessibility, Model-as-a-Service (MaaS) \cite{gan2023model} has served people and vital tools and drivers of innovation across various applications, including intelligent customer service \cite{liu2023icsberts}, healthcare \cite{tarcar2019healthcare}, finance \cite{wu2023bloomberggpt}, law \cite{lai2023large}, robotics \cite{zeng2023large}, investment \cite{gupta2023gpt}, education and training \cite{kasneci2023chatgpt}, artistic creativity \cite{roemmele2018automated}, and so on. As a result, the widespread use of LLMs in the medical field holds immense potential for providing intelligent support and assistance to improve healthcare quality and efficiency.

Fortunately, LLMs can effectively tackle the challenges of medical information. Firstly, by leveraging deep learning \cite{lecun2015deep} and NLP \cite{chowdhary2020natural} techniques, LLMs can deeply understand and semantically reason with medical texts. They can comprehend medical terminology, contextual relationships, and semantic structures, thereby enabling more accurate retrieval and processing of medical information. Secondly, LLMs can integrate diverse medical data sources, including medical literature, clinical guidelines, and case reports, offering comprehensive and multifaceted information. They can extract knowledge and insights from vast datasets, providing healthcare professionals and patients with more comprehensive and precise information support. Thirdly, LLMs can actively track medical literature and the latest research advancements, promptly delivering up-to-date information to healthcare professionals and patients. Furthermore, they can provide personalized recommendations and advice tailored to user needs and preferences, augmenting the relevance and practicality of the information. Therefore, researching LLMs in the medical field is necessary and crucial.

The process of training a medical LLM typically involves seven steps \cite{huang2023chatgpt}: data collection, data preprocessing, model selection and architecture design, model training, hyperparameter tuning, validation and evaluation, and model deployment and application. During the data collection phase, a large-scale corpus of text data in the medical domain is gathered, encompassing medical literature, case reports, clinical guidelines, drug information, and more. The collected data is then preprocessed, and an appropriate model architecture is selected and designed to handle the training and processing of the medical data. The model is trained, and its parameters are iteratively optimized to enhance its proficiency in the specialized domain. Finally, the model's performance is evaluated using a validation set employing metrics such as perplexity, text quality, and accuracy. Once the model training and evaluation are complete, they can be deployed in practical applications to fulfill the information needs of medical professionals and patients. By utilizing LLMs trained with specialized data inputs and multiple training iterations, these models acquire professional judgment capabilities. As a result, LLMs can meet the demands of medical professionals and patients for accurate, timely, and reliable medical information, thereby improving the quality and efficiency of medical decision-making. For instance, consider a scenario where a doctor faces a rare case and is uncertain about the diagnosis and treatment options. The doctor can employ an LLM to quickly search for relevant medical literature, case reports, and expert opinions, gaining more insights and guidance. Leveraging the vast pre-training data and semantic understanding capabilities of LLM, the model assists the doctor in comprehending the case's characteristics, pathophysiology, and potential treatment options. Moreover, the medical LLM can provide the doctor with assessments of the most recent clinical guidelines, drug information \cite{goel2023llms}, and treatment plans \cite{wilhelm2023large}, facilitating informed decision-making. Similarly, for patients, suppose an individual has been diagnosed with a rare disease and desires to acquire more information about the condition, treatment options, and lifestyle recommendations. The patient can utilize one LLM to input relevant keywords or questions, enabling them to access medical knowledge and research findings. The model searches for and presents the latest research discoveries, expert opinions, and information published by authoritative organizations, aiding the patient's understanding of the disease's symptoms, diagnostic methods, treatment options, and prognosis. Through interaction with the LLM, the patient can obtain easily understandable and accurate explanations, enhancing their comprehension of their condition and enabling more accurate communication with healthcare providers regarding their symptoms. Additionally, this empowers them to adopt scientifically sound treatment plans. Numerous medical LLMs have been exploited and are expected to apply to life and production \cite{minssen2023challenges}.

Therefore, exploring the applications, limitations, and potential advancements of LLMs in the medical domain is crucial. This up-to-date survey aims to provide a comprehensive overview of the utilization of LLMs in medicine, including their benefits, challenges, and emerging trends. By analyzing existing literature and research, we seek to offer insights into the current state of LLMs in medicine, compare different approaches, and identify areas for future research and development. To refine the contributions of this paper and provide more insights for further studies, we compare the similarities and differences between this paper and other review articles on the same subject, shown in Table \ref{table:contribution}, which makes our contribution clearer. This paper's contributions can be summarized as follows:

\begin{itemize}
    \item Comprehensive coverage. We provide an up-to-date and exhaustive survey of medical LLMs, including the advances in theory, methods, and applications.
    
    \item Progressive review. We reviewed the development stages of LLM, manifesting the respective advantages and disadvantages.
    
    \item Novel classification. According to various application fields, some representative medical LLM products are introduced, along with their training framework and process.
    
    \item Comprehensive discussion. We extensively explored the current trends in the medical grand model, delving into the opportunities and future directions it presents to assist in subsequent efforts in related domains.
\end{itemize}

\begin{figure}[ht]
    \centering
    \includegraphics[clip,scale=0.48]{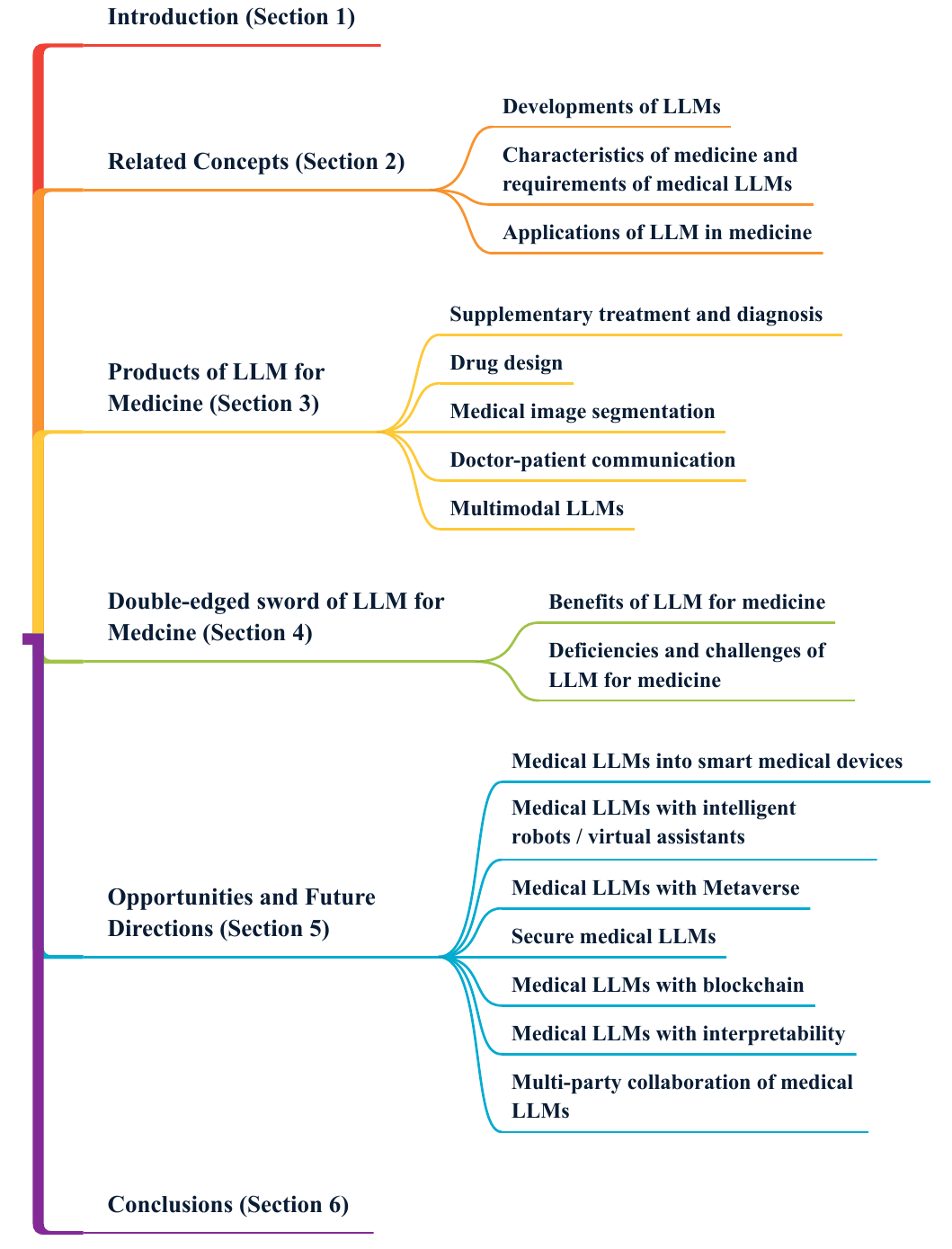}
    \caption{The outline of this article.}
    \label{fig:outline}
\end{figure}

\textbf{Organization}: The arrangements for this paper are as follows: In Section \ref{sec:relatedwork}, we review the history of LLM and discuss applications of the medical LLM. We make a comparison of the different products of MedGPT in Section \ref{Products}. Furthermore, we demonstrate the duality of the LLM in medicine in Section \ref{Double-edged}. We highlight the opportunities and provide promising directions in Section \ref{sec:oppotrunities}. Finally, we conclude this paper in Section \ref{sec:conclusion}. The outline of this article is shown in Figure \ref{fig:outline}. Moreover, in this paper, many acronyms are used to represent concepts, medical terms, types of models, and frequently studied models. Table \ref{table:acronyms} provides the most common and important terms used in our paper.

\begin{table*}[ht]
    \centering
    \footnotesize
    \caption{Contributions and gaps of existing papers.}
    \label{table:contribution}
    \begin{tabularx}{\textwidth}
    {|m{0.6cm}<{\raggedright}
    |m{0.6cm}<{\raggedright}
    |m{3.3cm}<{\raggedright}
    |m{3.6cm}<{\raggedright}
    |m{0.5cm}<{\raggedright}
    |m{3.6cm}<{\raggedright}
    |m{0.5cm}<{\raggedright}}
        \toprule
        \hline
        \multicolumn{1}{|c|}{\textbf{Ref.}} & \multicolumn{1}{c|}{\textbf{Year}} & \multicolumn{1}{c|}{\textbf{One-sentence summary}} & \multicolumn{1}{c|}{\textbf{Application of LLM}} & \multicolumn{1}{c|}{\textbf{Comparison}} & \multicolumn{1}{c|}{\textbf{Threat concerns}} & \multicolumn{1}{c|}{\textbf{Solution}}\\
        \hline
        \hline

        \cite{clusmann2023future} & 2023 & The future landscape of LLMs in medicine & \multicolumn{1}{c|}{\XSolidBrush}  & \multicolumn{1}{c|}{\XSolidBrush} & Ethics, training data security & \multicolumn{1}{c|}{\XSolidBrush} \\
        \hline
        
        \cite{thirunavukarasu2023large} & 2023 & LLMs in medicine & Chatbots & \multicolumn{1}{c|}{\Checkmark} & Recency, accuracy, coherence, interpretability, ethics & \multicolumn{1}{c|}{\Checkmark} \\
        \hline
        
        \cite{singhal2023large} & 2023 & LLMs encode clinical knowledge & \multicolumn{1}{c|}{\XSolidBrush} & \multicolumn{1}{c|}{\Checkmark} & 
         Evaluation data expansion, clinical accuracy, human evaluation framework, fairness, ethics & \multicolumn{1}{c|}{\XSolidBrush} \\
        \hline
        
        \cite{karabacak2023embracing} & 2023 & Embracing LLMs for medical applications: Opportunities and challenges & \multicolumn{1}{c|}{\XSolidBrush} & \multicolumn{1}{c|}{\XSolidBrush} & Central role, accuracy, cost, ethics, data privacy, regulatory framework & \multicolumn{1}{c|}{$\sqrt{}\mkern-9mu{\smallsetminus}$} \\
        \hline

        \cite{zhou2023survey} & 2023 & A survey of LLMs in medicine: Progress, application, and challenge & Health Support, Report Generation, Medical robotics, Translation, Education
         & \multicolumn{1}{c|}{\Checkmark} & Evaluation benchmarks, domain data limitations, new knowledge adaptation, behaviour alignment, ethics & \multicolumn{1}{c|}{\Checkmark} \\
        \hline

        \cite{kim2023chatgpt} & 2023 & Chatgpt and LLM chatbots: The current state of acceptability and a proposal for guidelines on utilization in academic medicine & Academic medicine &
        \multicolumn{1}{c|}{\Checkmark}  & Accuracy, ethics & \multicolumn{1}{c|}{\XSolidBrush} \\
        \hline

        Our work & 2024 & LLMs in medicine: A survey & Supplementary diagnosis, drug design, medical image segmentation, doctor-patient communication, multimodal & \multicolumn{1}{c|}{\Checkmark} & Computational resources, model efficiency, data-related, practical applications, fairness, privacy, accountability, patient autonomy & \multicolumn{1}{c|}{\Checkmark} \\
        \hline
        \hline
        \end{tabularx}
\end{table*}

\begin{table}[ht]
    \centering
    \caption{Important terms of acronyms and corresponding full form.}
    \label{table:acronyms}
    \begin{tabular}{|m{1.5cm}<{\raggedright}|m{5.3cm}<{\raggedright}|}
        \hline
        \multicolumn{1}{|c|}{\textbf{Acronym}} & \multicolumn{1}{c|}{\textbf{Full Form}} \\  
        \hline
        LLM & Large Language Model \\
        \hline
        RNN & Recurrent Neural Network \\
        \hline
        HMM & Hidden Markov Model \\
        \hline
        LSTM & Long Short-term Memory Network \\
        \hline
        ELMo & Embeddings from Language Model \\
        \hline
        GPT & Generative Pre-trained Transformer \\
        \hline
        BERT &Bidirectional Encoder Representations from Transformer \\
        \hline
        CBOW & Continuous Bag of Words Model \\
        \hline
        GRU & Gated Recurrent Unit \\
        \hline
        Med-PaLM & Medical Prompt Language Model \\
        \hline
        MSA & Multiple Sequence Alignment \\
        \hline
        PLM & Protein Language Model \\
        \hline
        WCE & Wireless Capsule Endoscopy \\
        \hline
        CFG & Category-Guided Feature Generation Module \\
        \hline
        MedLSAM & Medical Localize and Segment Anything Model \\
        \hline
        MLLM & Multimodal Large Language Model \\
        \hline
        EHR & Electronic Health Record \\
        \hline         
        \end{tabular}
\end{table}

\section{Related Concepts} \label{sec:relatedwork}
\subsection{Developments of LLMs}

\begin{figure}[ht]
    \centering
    \includegraphics[clip,scale=0.15]{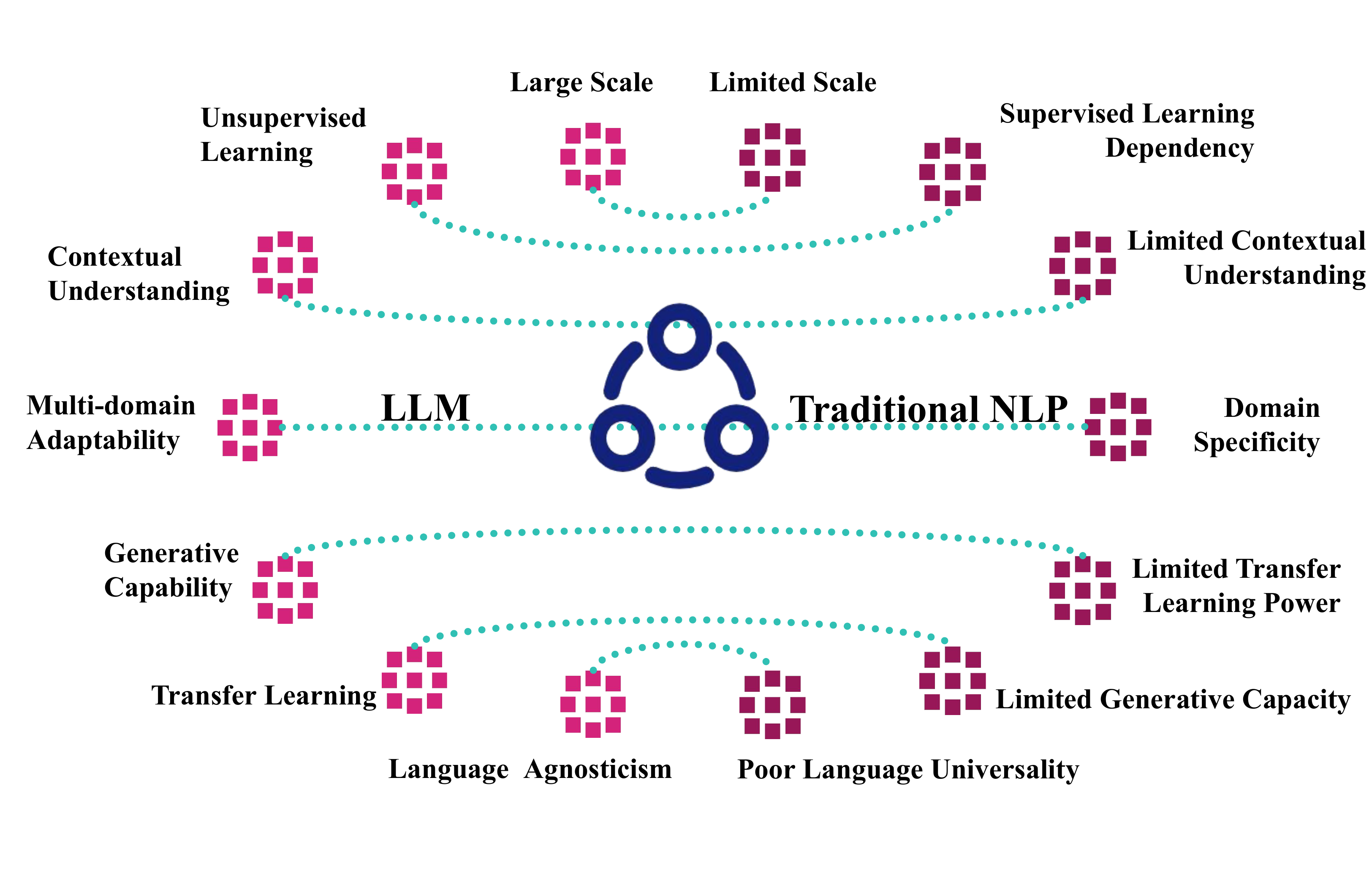}
    \caption{Characteristics of LLM vs. the traditional NLP.}
    \label{fig: Characteristics}
\end{figure}

The emergence of LLMs was not a straightforward process but rather a winding and progressive journey. In the early stages, studies focused on developing generative models capable of generating text \cite{hu2017toward}, images \cite{sarkar2021humangan}, audio \cite{kim2019flowavenet}, and other AI-generated content \cite{wu2023ai}. However, the performance of these models was limited due to the lack of large-scale datasets and powerful computational resources. Later, with the advent of pre-training models \cite{han2021pre,zeng2023distributed}, several studies began utilizing large-scale datasets for pre-training to extract statistical patterns and semantic representations from the data. The model could acquire initial parameters through pre-training and then serve as a fine-tuning model \cite{howard2018universal}. The above approach laid the foundation for subsequent model training, while numerous technical challenges in training and optimization awaited resolution. Subsequently, recurrent neural networks (RNNs) \cite{grossberg2013recurrent} were introduced as the foundational models for text generation. Although these models performed well in generating short texts and small-scale data, they encountered issues such as handling long sequences, vanishing gradients, and low computational efficiency. Nevertheless, with advancements in deep learning and computational resources, LLMs started demonstrating their immense potential \cite{wu2023multimodal}. Particularly, the introduction of the Transformer architecture enabled models to better capture contextual information and semantic representations of text, leading to breakthroughs in tasks such as text generation, dialogue systems, and natural language understanding. The aforementioned progressive development can be summarized in three stages: generative models, pre-training models, and autoregressive models. Meanwhile, the characteristics of LLM compared to the traditional NLP technology \cite{cambria2014jumping} are shown in Figure \ref{fig: Characteristics}.

\subsubsection{Generative models}

Generative models \cite{harshvardhan2020comprehensive} are probabilistic models that aim to generate new samples by generating a probability distribution that is similar to the original distribution. Through such a learning process, the model could generate new samples similar to the statistical patterns and patterns learned from large-scale datasets. In generative models, language models play an important role. The goal of a language model is to estimate the probability of a given word or a given sequence of words by training the model to learn the probability distribution of lexical representation in the sequences. This allows the model to predict the next word based on the previously observed words, thereby generating coherent text. Generative models generate new samples by computing the probability distribution of all the data. There are numerous typical models, such as N-gram models \cite{brown1992class}, hidden Markov models (HMMs) \cite{blunsom2004hidden}, and long short-term memory networks (LSTMs) \cite{yu2019review}.

\textbf{The N-gram model \cite{brown1992class}} has a relatively simple model structure and computational approach. Based on the previous $n$-1 words, the $N$-gram model predicts the next word, making it effective for short text generation and simple language modeling tasks. However, the $N$-gram model only considers the preceding $n$-1 words, failing to capture longer contextual information. Moreover, in cases of sparse data, the model may encounter problems of missing data and inefficient estimation. As a result, the $N$-gram model cannot capture long-distance dependencies and is sensitive to data sparsity issues.

\textbf{HMMs \cite{blunsom2004hidden}} are statistical models for defining transition and emission probabilities between states and observations. Because of the Markov chain structure \cite{tweedie2001markov} and conditional independence hypothesis, HMMs can effectively model the probability distribution of sequence data. Thus, they perform well in the tasks of labeling \cite{qiao2015diversified} and sequence generation \cite{kall2005hmm}. However, due to the Markov property for state transitions, they ignore longer contextual information. Also, they are limited and even suffer from problems with overfitting and data drift when the distributional assumptions of the training data are not strictly met. This happens when the distribution of the training data is different from the distribution of the data used in real life.


\textbf{LSTMs \cite{yu2019review}} are a special type of RNN that is used for solving the problems of gradient disappearance and gradient explosion \cite{philipp2017exploding} during long sequence training. Gate mechanisms remember important information and forget unimportant information in sequence data. They are suitable for processing sequence data \cite{lippi2019natural}, generating text, and capturing long-distance dependencies. However, they also have limitations. For example, when dealing with more complex language modeling tasks, both larger-scale data and computational resources are required due to the shortage of parallel processing.

\subsubsection{Pre-training models}

The pre-training models \cite{zeng2023distributed} aim to address the challenges in the generative models, such as generating text with semantic coherence, generating reasonable text, and capturing semantic relationships and information in context. Pre-training models are based on neural networks. They extract semantic representations of text from large-scale corpora by learning language knowledge. This approach enables the model to have a better understanding of the data and extract features. Common pre-training models include Word2Vec \cite{church2017word2vec}, Embeddings from Language Models (ELMo) \cite{ethayarajh2019contextual}, Generative Pre-trained Transformer (GPT) \cite{roumeliotis2023chatgpt}, Bidirectional Encoder Representations from Transformers (BERT) \cite{kenton2019bert}, and others. Note that the pre-training model has broader applications and can be used for various tasks such as word embedding \cite{luo2014pre}, text classification \cite{zheng2020pre}, named entity recognition \cite{yohannes2022named}, data mining \cite{gan2017data,huang2023us}, and more.

\textbf{Word2Vec \cite{church2017word2vec}} is an early milestone in pre-training models, as proposed in 2013. It is a pre-training model based on the distributed hypothesis. Word2Vec utilizes two training methods: the Continuous Bag of Words (CBOW) model \cite{wang2017two} and the Skip-gram model \cite{mccormick2016word2vec}. The CBOW model aims to predict the center word based on the surrounding context words, while the skip-gram model predicts the context words given the center word. A neural network model is trained to map words to fixed-dimensional representations in a continuous vector space. This representation allows for capturing the semantic information of words and calculating the similarity between words. However, Word2Vec can only represent the semantics of individual words but cannot capture contextual information or handle the semantics of polysemous words.

\textbf{ELMo \cite{ethayarajh2019contextual}} is a pre-training model proposed by the Allen Institute for Artificial Intelligence in 2018. It employs a bi-directional language model to generate vector representations of words. During the pre-training process, ELMo trains two language models simultaneously, one from left to right and the other from right to left, to acquire contextually relevant word representations. Ultimately, ELMo combines these representations to form the final vector representation for each word. This context-sensitive representation enables ELMo to better capture word polysemy and contextual relevance. However, ELMo, based on bi-directional language models, has slower training and inference processes and cannot handle sentence-level semantic relationships.

\textbf{GPT \cite{roumeliotis2023chatgpt}} was first introduced by OpenAI in 2018 and has been continuously improved in subsequent versions. It utilizes the transformer model structure \cite{wolf2020transformers} and is trained in an autoregressive manner. GPT constructs its model by stacking multiple transformer encoder layers, where each layer can leverage the contextual information from the preceding text. Through large-scale unsupervised pre-training, it learns language models that can be fine-tuned for various downstream tasks, such as text classification and machine translation. GPT aims to address text generation and language understanding challenges. GPT is capable of generating coherent text and comprehending contextual information when trained in an autoregressive manner. However, GPT still has room for improvement as it lacks precise control over input sentences, and the generated results may occasionally lack accuracy.

\textbf{BERT \cite{kenton2019bert}} is a pre-training model introduced by Google in 2018. BERT uses a bidirectional transformer model, which simultaneously predicts both left and right words based on context. Unlike GPT, BERT takes into account the contextual information from both the left and right sides. During the pre-training process, BERT predicts the masked words by masking a portion of the input words. Furthermore, BERT incorporates sentence-level pre-training tasks, such as predicting sentence continuity. This bidirectional training approach allows BERT to have a better understanding of contextual information and handle polysemous words and ambiguous sentences, thereby providing more accurate semantic representations. However, BERT cannot directly handle text-generation tasks.

\subsubsection{Autoregressive models}

The autoregressive model is part of pre-training models, although they differ in their objectives. The autoregressive model aims to generate coherent text by predicting the next word or character based on context information. In contrast, the objective of the pre-training model is to learn language representations for fine-tuning or transfer learning in subsequent tasks. The autoregressive model is trained through supervised learning and requires a large amount of labeled data. Conversely, the pre-training model is unsupervised and utilizes large-scale unlabeled text data for pre-training, without specific annotations. The autoregressive model excels in text generation tasks and is commonly used in natural language generation \cite{zhang2023tractable}, and similar tasks.

In the autoregressive model, text generation is accomplished by sequentially predicting the next word or character. The model generates the next element in the sequence using the current context information and adds it to the generated sequence. This recursive generation approach enables the model to maintain contextual coherence. Representative models in this stage include RNN and its variants, such as LSTM and gated recurrent units (GRU) \cite{dey2017gate}. On the other hand, the pre-training model GPT is based on the transformer architecture and utilizes autoregressive training. This model combines the advantages of both the autoregressive model and the pre-training model.

\textbf{Rerrent Neural Network (RNN) \cite{mikolov2010recurrent}} is one of the earliest autoregressive models applied to natural language processing tasks. RNN can handle variable-length sequential data and capture contextual information within the sequence. It has achieved certain accomplishments in language modeling and sequence generation tasks. The basic idea is to capture contextual information within the sequence by recursively passing hidden states. However, traditional RNNs suffer from the problem of vanishing or exploding gradients when dealing with long-term dependencies.

\textbf{Gate Recurrent Unit (GRU) \cite{dey2017gate}}  is another improved RNN structure that simplifies LSTM's gating mechanism. Compared to LSTM, GRU has fewer parameters, higher computational efficiency, and comparable performance on certain tasks. However, GRU still cannot fully address the issue of long-term dependencies and may perform poorly on tasks that require handling long-distance dependencies.

\textbf{Transformer \cite{han2021transformer}} is a model that uses a self-attention mechanism and introduces the encoder-decoder structure for sequence-to-sequence tasks. The transformer does not rely on recursive structures and can be computed in parallel, considering all positions in the input sequence simultaneously, regardless of sequence length. Therefore, it is more efficient and effective in handling long texts and modeling long-distance dependencies. However, transformer also suffers from the drawback of higher computational complexity, especially when dealing with longer sequences, which consume significant computational resources. Additionally, the transformer is sensitive to positional information within the sequence and may require additional encoding to retain positional information. GPT is an autoregressive model based on the Transformer architecture.

\subsection{Characteristics of medicine and requirements of medical LLMs}

\textbf{Compassionate care.} Medicine emphasizes compassionate care for patients \cite{tehranineshat2019compassionate}, including establishing trust, respecting patient rights and dignity, and addressing patients' psychological and social aspects. Therefore, when interacting with patients, medical LLMs need to embody characteristics of compassionate care, such as respecting patient privacy and addressing emotional and social needs.

\textbf{Interpretability.} Medicine encompasses a broad range of knowledge, and medical decisions are made based on this knowledge. Given the critical nature of medical decision-making, results from medical LLMs need to possess good interpretability \cite{vellido2020importance}. This ensures that both healthcare professionals and patients can understand the model's reasoning process and conclusions, thereby enhancing trust.

\textbf{Practice-oriented.} Medicine is a practice-oriented discipline \cite{zink2004invasive}, where the application of medical knowledge is typically validated through clinical practice. To ensure the models' practicality and applicability, the design and application of medical LLMs must align with medical practices, by taking into account real-world clinical scenarios and healthcare services.

\textbf{Team collaboration.} Collaboration within healthcare teams is essential in the field of medicine \cite{pollack2012emergency}. Various healthcare professionals, including doctors, nurses, laboratory technicians, and rehabilitation specialists, need to work together to provide comprehensive healthcare services. Recognizing the collaborative nature of the medical field, medical LLMs should be capable of collaborating with other healthcare professionals, supporting integrated decision-making and services within multidisciplinary healthcare teams.

\textbf{Ethical challenges.} Medicine involves numerous ethical considerations, encompassing issues such as patient privacy, bioethics, and treatment decisions. Healthcare professionals must confront and address these ethical challenges to ensure fairness and ethical practices in medical care. Similarly, the design and application of medical LLMs should adhere to medical ethical principles \cite{muller2021ten}, emphasizing patient privacy and ensuring fairness and ethical decision-making processes.

\textbf{Uncertainty and complexity.} The human body is an incredibly complex system, with numerous factors influencing health and disease, leading to inherent uncertainty in medical practice. When faced with complex cases, medical professionals must make comprehensive judgments \cite{bhise2018defining}. Therefore, medical LLMs should effectively handle this uncertainty, providing probabilistic results and decision-making capabilities.

\textbf{Diverse fields.} Medicine encompasses a variety of specialized domains, including internal medicine, surgery, obstetrics, and gynecology, among others. Medical LLMs need to provide tailored support based on the specialized knowledge of different fields, thereby assisting doctors in making accurate diagnoses and treatment decisions within specific domains. Medicine involves addressing a variety of ailments and causes, including genetic diseases, infectious diseases, chronic conditions, and more. Medical LLMs must offer personalized medical recommendations for different ailments and causes, enhancing the specificity of patient treatment plans. The medical process encompasses multiple stages, including prevention, diagnosis, treatment, and rehabilitation. Medical LLMs need to provide support at each stage, such as formulating personalized prevention plans, aiding in clinical decision-making, and offering rehabilitation advice, thereby comprehensively serving patients throughout their entire medical journey. As shown in Table \ref{table: support}, medical LLMs can provide assistance tailored to different medical fields.

\begin{table*}[ht]
    \centering
    \caption{Various medical fields and LLMs support}
    \label{table: support}
    \begin{tabular}{|m{3.2cm}<{\raggedright}|m{3.3cm}<{\raggedright}|m{3.3cm}<{\raggedright}|m{5.2cm}<{\raggedright}|}
        \hline
        \hline
        \multicolumn{1}{|c|}{\textbf{Medical Field}} & \multicolumn{1}{c|}{\textbf{Patient Characteristics}} & \multicolumn{1}{c|}{\textbf{Typical Disease Types}} & \multicolumn{1}{c|}{\textbf{Support by Medical LLMs}} \\
        \hline
        \hline
        Internal medicine \cite{omiye2024large} & Age, family medical history & Diabetes, hypertension, cardiovascular diseases & Formulate personalized prevention plans, assist in clinical decision-making, provide patient rehabilitation advice, and generate drug interaction assessments \\
        \hline
        Surgery \cite{puladi2023impact} & Surgical history, trauma history & Surgical procedures, trauma & Assist in pre- and post-surgery treatment plans, offer rehabilitation advice, and provide real-time surgical guidance \\
        \hline
        Obstetrics and gynecology \cite{grunebaum2023exciting} & Women's age, obstetric and reproductive history & Pregnancy management, gynecological diseases & Provide pregnancy health advice, assist in the diagnosis and treatment of gynecological diseases and generate personalized contraception recommendations \\
        \hline
        Infectious diseases \cite{schwartz2023black} & Immunization status, travel history & Infectious diseases, bacterial infections & Develop infection prevention strategies, assist in diagnosing infectious cases and generate tailored antibiotic prescriptions \\
        \hline
        Genetic medicine \cite{feldman2019development} & Family genetic history, genetic information & Genetic diseases & Provide genetic diagnosis and treatment recommendations for genetic diseases, and offer personalized genetic counselling \\
        \hline
        Chronic diseases \cite{biswas2023role} & Lifestyle, long-term medication history & Hypertension, diabetes, chronic obstructive pulmonary disease & Develop long-term treatment plans, provide advice on managing chronic diseases and offer personalized dietary and exercise recommendations \\
        \hline
        \hline
    \end{tabular}
\end{table*}

\subsection{Applications of LLM in medicine}

Medical LLM exhibits potential capability in various fields, including dentistry \cite{huang2023chatgpt}, radiology \cite{akinci2023large}, nuclear \cite{alberts2023large}, clinical \cite{singhal2023large}, and drug design \cite{chakraborty2023artificial}. It employs common effects and unique functions in different fields. Generally speaking, medical LLMs can manage medical records and medical documents automatically. Compared to the traditional training model, the medical LLM showed advantages in training data, such as BioBERT \cite{lee2020biobert}, BioGPT \cite{luo2022biogpt}, etc. To illustrate in detail, take BioBERT as an example. BioBERT, pre-trained on a vast biomedical corpus, exhibits exceptional adaptability to domain-specific contexts and excels in comprehending and processing intricate biomedical terminologies and entities. Its performance surpasses that of other models in various biological and medical text mining tasks, particularly in three key areas: First, BioBERT demonstrates a 0.62\% improvement in the F1 score for recognizing entities within biomedical texts, including genes, proteins, and diseases, compared to alternative models. Secondly, it achieves a noteworthy 2.80\% enhancement in the F1 score for extracting relationships between entities from biomedical texts relative to other models. Thirdly, BioBERT exhibits a substantial 12.24\% improvement in Mean Reciprocal Rank for answering biomedical-related queries compared to its counterparts. Additionally, BioBERT's structural similarity to the widely used BERT model makes it easier to use and allows for smooth integration into a wide range of biological and medical text-mining tasks. As a result, it can be inferred that the LLM possesses a commendable degree of generalization capability on different data.

\begin{figure*}[ht]
    \centering
    \includegraphics[clip,scale=0.3]{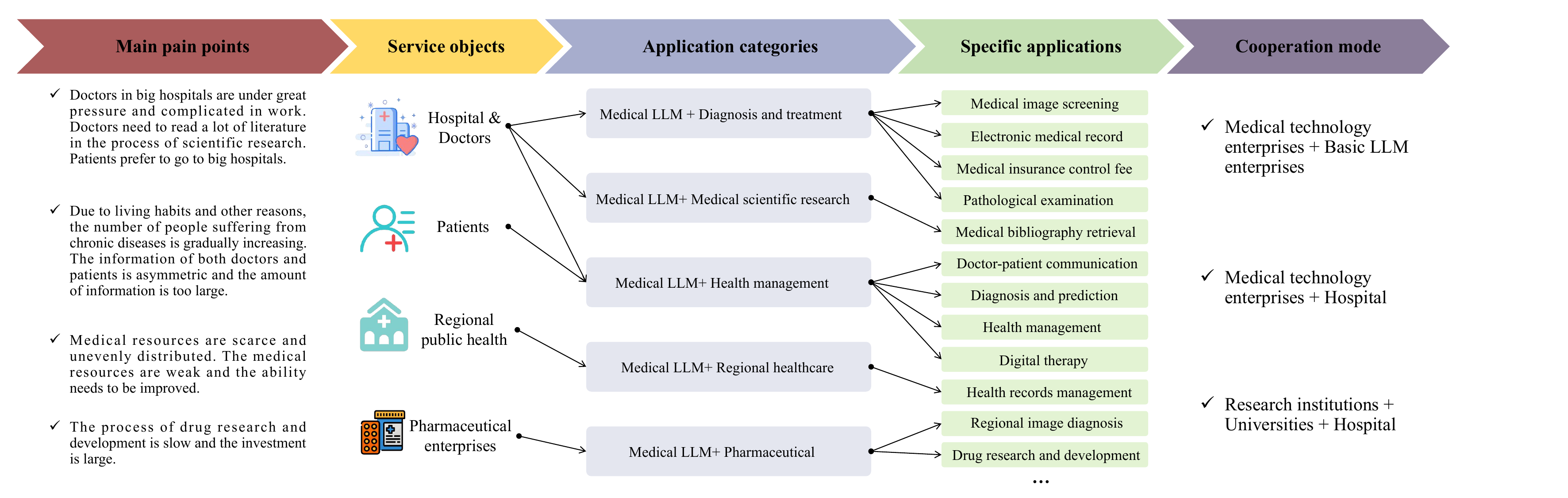}
    \caption{The pain points of medical treatment and the applications of medical LLMs.}
    \label{fig: applications}
\end{figure*}

For example, in dentistry, there are two main LLM deployment methods: automated dental diagnosis and cross-modal dental diagnosis \cite{huang2023chatgpt}. It points out how a multi-modal LLM AI system works for dentistry clinical application and exhibits the whole process. It can handle unstructured data, and extract and integrate information for doctors and patients. Furthermore, multiple documents assist medical LLM in efficiently providing treatment methods based on patients' backgrounds. With access to abundant professional training data and the ability to capture context information, medical LLMs aid in diagnosis. Not only does LLM have the ability to read text content, but it can also give interpretations for images. For instance, in radiology \cite{jeblick2023chatgpt}, medical LLM strengthens communications between doctors and patients by generating simplified reports. These reports are generally considered correct by most participating radiologists. Furthermore, medical LLM can analyze medical image reports, providing deeper interpretation and background knowledge to assist doctors in diagnosing diseases more accurately \cite{lecler2023revolutionizing}. Recent research has shown that medical LLMs, trained with the latest data, can provide clinical decision support based on the latest research findings, thereby benefiting clinical medicine. If combined with image interpretations, medical LLM may apply to clinical nuclear medicine to provide comprehensive information on imaging results, similar to clinical radiology \cite{shaikh2021artificial}. In addition, the application of medical LLMs in drug design can accelerate the process of drug target discovery and help predict different characteristics of drugs \cite{chakraborty2023artificial}. The applications of LLM in medicine are shown in Figure \ref{fig: applications}.

\section{Products of LLM for Medicine} \label{Products}

According to incomplete statistics, since 2023, the number of domestic and international large AI models released in the medical field has exceeded 30, and their application scenarios cover many aspects such as medical scientific research, drug research and development, smart diagnosis and treatment, medical equipment operation and maintenance, and hospital management. Except for the pre-trained data, the medical LLM also exhibits excellent capability of generalizability reflecting its outstanding experiment results in various application fields.
In the following, some well-known LLM products for medicine are summarized in detail, the specific applications of various LLMs in each link are illustrated in Figure \ref{fig: LLM_application_of_each_link}:

\begin{figure*}[ht]
    \centering
    \includegraphics[clip,scale=0.5]{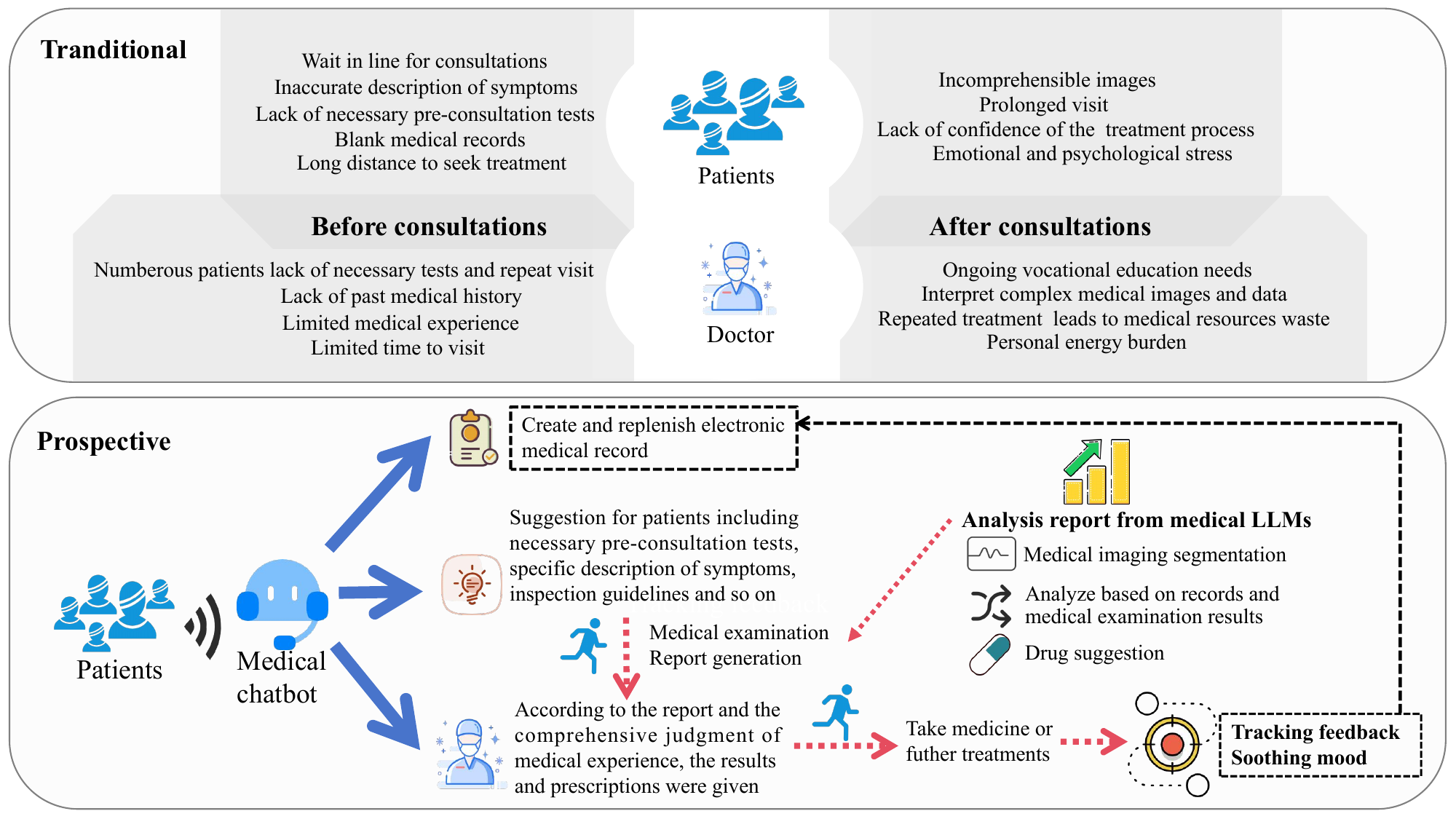}
    \caption{{Applications of LLMs in each link in medicine}}
    \label{fig: LLM_application_of_each_link}
\end{figure*}

\subsection{Supplementary treatment and diagnosis}

\textbf{BenTsao (Huatuo) \cite{wang2023huatuo}.} BenTsao, originally named Huatuo GPT, was trained by the research team at the Harbin Institute of Technology. BenTsao is a large-scale Chinese language model in the biomedical field, built upon the open-source LLaMa-7B model \cite{zeng2022glm}. The BenTsao model incorporates structured and unstructured medical knowledge from the Chinese Medical Knowledge Graph (CMeKG), which provides medical information about diseases, drugs, symptoms, and more. The CMeKG includes over 10,000 diseases, nearly 20,000 drugs, over 10,000 symptoms, and structured knowledge descriptions for 3,000 diagnostic and therapeutic techniques. This allows for extensive knowledge associations between diseases, symptoms, drugs, and diagnostic and therapeutic techniques, with 1.56 million concept relationships and attribute triplets. The BenTsao team samples instances from the knowledge graph based on specific task knowledge and uses the OpenAI API (GPT 3.5) \cite{koubaa2023gpt} to construct question-answer data around the medical knowledge base. They employ various prompt forms to fully utilize the knowledge, creating over 8,000 instruction data points for the instruction dataset used in supervised fine-tuning. For medical question-answering tasks, BenTsao introduces a new evaluation metric, the system usability scale (SUS). The SUS metric consists of three dimensions: safety, usability, and smoothness. Safety evaluates whether the generated responses could mislead users and pose a threat to their health. Usability evaluates the degree to which the generated responses reflect medical expertise, while smoothness measures the fluency of the generated responses.

\textbf{Med-PaLM \cite{singhal2022large}.} In July 2023, Google unveiled a new evaluation benchmark for its fine-tuned medical prompt language model (Med-PaLM), known as MultiMedQA, aimed at assessing the clinical capabilities of LLMs in the field of medicine. This benchmark encompasses questions and answers from various domains, such as medical exams and research, intending to test the model's performance in handling clinically relevant issues. Based on prompt-based fine-tuning, Med-PaLM's responses were compared to consumer medical queries with those of clinical professionals within a human-assessed framework. The results demonstrated outstanding performance by Med-PaLM, confirming the effectiveness of prompt-based fine-tuning. Particularly noteworthy is Med-PaLM's achievement on the MedQA dataset, where it surpassed the "passing" score in the style of U.S. Medical Licensing Examination (USMLE) questions for the first time, achieving a score of 67.2\%. However, the study highlighted that, despite significant progress, there is still room for improvement, especially when comparing model responses to those of clinical professionals. This suggests the potential for further enhancements in the quality of medical questions answered by the model. 

\subsection{Drug design}

\textbf{PanGu drug model \cite{lin2023pangu}.} It is a deep learning-based model designed for a variety of drug development applications. The inspiration for the model's design derives from the various representations of molecules that students learn in chemistry courses, such as molecular formulas and structural formulas. Moreover, it learns how to transform between these representations. By utilizing an asymmetric conditional variational autoencoder from graph to sequence, the PanGu drug model can effectively represent molecular features and enhance the performance of downstream drug discovery tasks. During the pre-training phase, the PanGu drug model was trained on 1.7 billion small molecules. Such extensive training allows the model to capture molecules' rich features and learn intricate relationships between them. In various drug discovery tasks, encompassing molecular property prediction, compound-target interactions, drug-drug interactions, chemical reaction yield prediction, molecular generation, and molecular optimization, the PanGu drug model has consistently achieved state-of-the-art results. Note that the PanGu drug model encompasses several distinctive functionalities. The PanGu molecule generator can generate novel compounds with analogous physicochemical properties, thereby expanding existing compound databases. These features are invaluable for drug research, providing a broader pool of candidate compounds for further research and screening. Additionally, the PanGu molecule optimizer can refine the chemical structure of initial molecules to enhance specific molecular properties, offering a potent tool for drug design and optimization. To enhance user accessibility, the PanGu drug model has developed an automated multi-objective optimization web application, the website shown in Table \ref{table:algorithm}.

\textbf{Large-scale protein language model (HelixFold-Single) \cite{fang2023method}.} Compared to traditional methods dependent on multiple sequence alignment (MSA), HelixFold-Single is an innovative protein structure prediction method to overcome limitations and time constraints. The HelixFold-Single method mainly integrates a large-scale protein language model (PLM) with key components derived from AlphaFold2. Initially, self-supervised learning was employed to learn an approach to pre-train the model on the original structures of billions of proteins and create a robust protein language model. This pre-trained model serves as a substitute for conventional MSA. It enables the acquisition of shared evolutionary information from protein sequences. Subsequently, the pre-trained PLM was amalgamated with several pivotal components of AlphaFold2 to form an end-to-end differentiable model. This model can predict the three-dimensional coordinates of a protein's native structure, thereby facilitating protein structure prediction. In contrast to traditional methodologies, the HelixFold-Single approach obviates the need for time-intensive MSA searches, thereby conferring significant advantages in terms of time efficiency. Furthermore, through validation on the CASP14 and CAMEO datasets, as for targets characterized by a substantial number of homologous protein families, the HelixFold-Single method has been demonstrated to achieve accuracy comparable to MSA-based methods. Additionally, the method exhibits shorter runtimes for tasks requiring numerous predictions, underscoring its potential applicability in high-throughput scenarios.

\subsection{Medical image segmentation}

\textbf{Deep Synergistic Interaction Network (DSI-Net) \cite{zhu2021dsi}.} DSI-Net is a deep learning approach designed for computer-aided diagnosis systems for gastrointestinal diseases, focusing on analyzing Wireless Capsule Endoscopy (WCE) images. In this system, WCE images are utilized to assist doctors in detecting and diagnosing lesions in patients. Traditional methods typically handle image classification and segmentation tasks separately, neglecting their interrelatedness and complementarity. Based on the backbone network DeeplabV3+, DSI-Net enhances overall performance by maximally exploiting the information exchange between these two tasks through the joint use of the classification branch, coarse segmentation branch, and fine segmentation branch. In the classification task, DSI-Net introduces the Lesion Location Mining module, which accurately highlights the lesion area to improve classification results. It enhances lesion detection and localization accuracy by mining ignored lesion areas and erasing background regions erroneously classified. For segmentation tasks, DSI-Net proposes the Category-Guided Feature Generation module (CFG), which utilizes category prototypes learned in the classification task to improve pixel representations, resulting in more accurate segmentation outcomes. By leveraging category information from the classification branch, the CFG module can generate features relevant to specific categories, thereby enhancing segmentation accuracy. Additionally, DSI-Net incorporates a task interaction loss to strengthen mutual supervision between classification and segmentation tasks, ensuring consistency in their predicted results. Through this approach, DSI-Net effectively leverages the interdependencies between classification and segmentation tasks to enhance overall performance.

\textbf{Medical localize and segment anything model (MedLSAM) \cite{lei2023medlsam}.} MedLSAM primarily focuses on localization and segmentation tasks in the field of medical image analysis. Traditional image segmentation methods often require layer-by-layer annotation, which is time-consuming and challenging when dealing with large-scale medical image datasets. MedLSAM aims to address this issue. The MedLSAM model's core idea is to combine localization and segmentation tasks, resulting in automated segmentation through self-supervised learning. It introduces a 3D localization base model called MedLAM, which can accurately locate target anatomical structures in the images. To train the MedLAM model, a dataset containing a large number of CT scan images is used, and training is conducted through self-supervised learning tasks. By performing extreme point annotation on a small number of templates, the MedLSAM model can automatically identify and locate the target anatomical regions in unlabeled data. Once the localization task is completed, the MedLSAM model generates 2D bounding boxes, which are then used for precise segmentation with existing image segmentation models. This combined localization and segmentation approach makes the segmentation process more efficient and accurate. Extensive experimental evaluations were conducted on two 3D datasets containing multiple different organs to assess the MedLSAM model. The experimental results demonstrate that the model achieves outstanding performance in both localization and segmentation tasks, with a lower dependency on extreme point annotation.

\subsection{Doctor-patient communication}

\textbf{PubMed GPT \cite{venigalla2022pubmed}.} It is a sophisticated LLM, tailored for the biomedical domain. It leverages its advanced natural language processing capabilities to facilitate a broad spectrum of academic research and clinical applications. The GPT-3.5 architecture was used to develop the PubMed GPT model, representing an enhanced iteration of the encoder-decoder structure based on the transformer model. This model encompasses multiple layers, each comprising self-attention mechanisms and feedforward neural networks. Within the encoder component, input text is processed through multiple layers of self-attention. The self-attention mechanism empowers the model to automatically assign weights to different positions in the input sequence, thereby capturing contextual information more effectively. In each self-attention layer, the model computes attention scores to determine the significance of each position to others. Using these scores, the model performs a weighted summation of the inputs, allowing for a more refined focus on contextual information relevant to the current position during input processing. In the decoder, a similar self-attention mechanism is applied to manage the output sequence. To preserve the sequence's order, the model incorporates positional encoding. This technique embeds positional information into the input sequence, which allows the model to discern between words at different positions. By combining self-attention mechanisms and positional encoding, the decoder can conscientiously consider the order and context of the input sequence during output generation. Significantly, the GPT-3.5 architecture is characterized by an extensive parameter count, boasting 175 billion trainable parameters \cite{koubaa2023gpt}. This expansive parameterization equips the model to adeptly capture intricate semantic and grammatical rules, delivering heightened proficiency in text generation and comprehension. Before training the PubMed GPT model, an extensive corpus of biomedical literature data was utilized as pre-training data. By pre-training on self-supervised tasks using this data, the model assimilated the semantics and grammatical rules inherent to the biomedical field, establishing a nuanced understanding of biomedical terms and concepts. This comprehensive training regimen endows the model with superior accuracy and domain-specific acumen for text-processing tasks within the biomedical realm.

\textbf{ChatDoctor \cite{yunxiang2023chatdoctor}.} ChatDoctor is a fine-tuned medical chat model designed to provide high-quality medical advice and guidance. The model is built upon the LLM for medicine and artificial intelligence (LLaMA) infrastructure and has been fine-tuned using a dataset of 100,000 patient-doctor dialogues from online medical consultation platforms. To ensure patient confidentiality, this dataset undergoes rigorous privacy protection and personal information cleaning. One of the ChatDoctor model's key innovations is the self-retrieval mechanism for information. This mechanism allows the model to retrieve and leverage real-time information from online resources like Wikipedia and offline medical databases. This capability enables the model to better comprehend patient queries and deliver accurate and reliable medical advice. By combining an LLM's powerful language understanding capabilities with real-time knowledge supplementation through self-retrieval, the ChatDoctor model can provide patients with more comprehensive and personalized medical consultations. Through fine-tuning with real-world patient-doctor interaction dialogue data, the ChatDoctor model has shown significant improvements in understanding patient needs and providing recommendations. When answering medical queries, the model's performance has undergone rigorous evaluation and comparison, demonstrating higher accuracy and reliability compared to other existing medical LLMs.

\begin{table*}[hb]
    \centering
	\caption{Information of different medical LLMs.}
	\label{table:algorithm}
    \footnotesize
    \renewcommand{\arraystretch}{1.15}
	\begin{tabularx}{\textwidth}{m{5cm}<{\centering}|m{2.5cm}<{\centering}|m{0.6cm}<{\centering}|m{0.7cm}<{\centering}|m{6.5cm}<{\raggedright}}
		\toprule
		\hline
		\textbf{Domain} & \textbf{LLM} & \textbf{Year} & \textbf{Paper} & \multicolumn{1}{c}{\textbf{Source}} \\  
        \hline

        \multirow{7}{*}{Supplementary treatment and diagnosis}
        & MedGPT & 2021 & \cite{kraljevic2021medgpt} & \url{https://medgpt.co/home} \\
        \cline{2-5}
        
        & LLM-Mini-CEX & 2023 & \cite{shi2023llm} & \multicolumn{1}{c}{$/$}  \\
        \cline{2-5}
        
        & WiNGPT & 2023 & $/$ & \url{https://github.com/winninghealth/WiNGPT2} \\ 

        \cline{2-5}
        
        & SkinGPT-4 & 2023 & \cite{zhou2023skingpt} & \multicolumn{1}{c}{$/$} \\

        \cline{2-5}

        & DoctorGLM & 2023 & \cite{xiong2023doctorglm} & \url{https://github.com/xionghonglin/DoctorGLM} \\ 
        \cline{2-5}

        & BenTsao (Huatuo) & 2023 & \cite{wang2023huatuo} & \url{https://github.com/SCIR-HI/Huatuo-Llama-Med-Chinese} \\
        \cline{2-5}

        & ClinicalGPT & 2023 & \cite{wang2023clinicalgpt} & \multicolumn{1}{c}{$/$} \\

        \hline

        \multirow{4}{*}{Drug design}
        & PanGu Drug Model & 2023 & \cite{lin2023pangu} & \url{http://www.pangu-drug.com/}  \\
        \cline{2-5}

        & HelixFold-Single & 2023 & \cite{fang2023method} & \url{https://github.com/PaddlePaddle/PaddleHelix/tree/dev/apps/protein_folding/helixfold-single} \\
        \cline{2-5}

        & TransAntivirus & 2023 & \cite{mao2023transformer} & \url{https://github.com/AspirinCode/TransAntivirus.} \\
        \cline{2-5}

        & OpenBioMed & 2023 & \cite{luo2023unified} & \url{https://github.com/PharMolix/OpenBioMed} \\ 
        
        \hline
        
        \multirow{4}{*}{Medical image segmentation}
        & DSI-Net & 2021 & \cite{zhu2021dsi} & \url{https://github.com/CityU-AIM-Group/DSI-Net}  \\
        \cline{2-5}
        
        & MedLSAM & 2023 & \cite{lei2023medlsam} & \url{https://github.com/openmedlab}  \\

        \cline{2-5}
        
        & Lvit & 2023 & \cite{li2023lvit} & \url{https://github.com/HUANGLIZI/LViT}  \\

        \cline{2-5}
        
        & MedCLIP-SAM & 2024 & \cite{koleilat2024medclip} & \multicolumn{1}{c}{$/$}  \\
        
        \hline
        
        \multirow{6}{*}{Doctor-patient communication} 
        & BioMedLM $/$ PubMed GPT & 2022 &\cite{venigalla2022pubmed} & \url{https://www.mosaicml.com/blog/introducing-pubmed-gpt} \\

        \cline{2-5}
        
        & ChatDoctor & 2023 & \cite{yunxiang2023chatdoctor} & \url{https://github.com/Kent0n-Li/ChatDoctor}  \\
        \cline{2-5}
        
        & Disc-medllm & 2023 & \cite{bao2023disc} & \url{https://github.com/FudanDISC/DISC-MedLLM} \\
        \cline{2-5}
        
        & BianQue & 2023 & \cite{chen2023bianque} & \url{https://github.com/scutcyr/BianQue} \\
        \cline{2-5}
        
        & MeChat & 2023 & \cite{Qiu2023SMILE} & \url{https://github.com/qiuhuachuan/smile} \\
        \cline{2-5}
        
        & PMC-LLaMA & 2023 & \cite{wu2024pmc} & \url{https://github.com/chaoyi-wu/PMC-LLaMA} \\

        \hline
        
        \multirow{3}{*}{Multimodal} 
        & OpenMEDLab & 2023 & $/$ & \url{https://github.com/openmedlab} \\
        \cline{2-5}
        
        & Med-MLLM & 2023 & \cite{liu2023medical} & \multicolumn{1}{c}{$/$}\\
        \cline{2-5}
        
        & PeFoMed & 2024 & \cite{he2024pefomed} & \url{https://github.com/jinlHe/PeFoMed}\\
        \hline

        \multirow{4}{*}{Health management}
        & CIDRS & 2021 & \cite{wang2021cloud} & \multicolumn{1}{c}{$/$}
        \\
        \cline{2-5}

        & GatorTron & 2022 & \cite{yang2022large} & \url{https://catalog.ngc.nvidia.com/orgs/nvidia/teams/clara/models/gatortron_og}
        \\
        \cline{2-5}

        & CareGPT  & 2023 & $/$ & \url{https://github.com/WangRongsheng/CareGPT}  \\
        \cline{2-5}
        
        & Bianshi & 2023 & $/$ & \url{https://www.a-eye.cn/technology.html##Model} \\
        \hline

        \bottomrule
        \end{tabularx}
\end{table*}

\subsection{Multimodal LLMs}

\textbf{Med-MLLM \cite{liu2023medical}.} It is a medical multimodal large language model (MLLM) designed to address the challenges posed by future pandemics. By learning extensive medical knowledge from unlabeled data, including image understanding, semantic text, and clinical phenotypes, the model can be rapidly deployed and adapted to rare diseases, such as emerging pandemics. The Med-MLLM framework supports the processing of visual modalities (chest X-rays, CT scans, etc.) and text modalities (medical reports, clinical notes, etc.) in medical data, making it applicable to clinical tasks that require simultaneous handling of both visual and textual data. The design objective of Med-MLLM is to overcome challenges faced by traditional neural network models in scenarios involving rare diseases, where there is a lack of sufficient labeled data. Due to the high cost and time-consuming nature of collecting and labeling data for rare diseases, traditional methods encounter difficulties. To address this issue, Med-MLLM employs large-scale pre-training techniques, shortening the model's deployment time and enabling rapid responses to the emergence of future rare diseases. The Med-MLLM framework employs a structured training approach, including pre-training image encoders and text encoders for handling visual and textual data. To further enhance the model's performance, a soft image-text alignment loss is introduced for pre-training visual and text encoders. This multimodal pre-training approach enables Med-MLLM to simultaneously process visual, textual, and multimodal inputs, demonstrating accurate and robust performance in tasks such as COVID-19 reporting, diagnosis, and prognosis. The study also conducted extensive evaluations of Med-MLLM, testing its performance on COVID-19 pandemic data and showcasing its accuracy and robustness in decision-support tasks with limited labeled data. Additionally, they studied 14 other common chest diseases and tuberculosis, finding that even with only 1\% labeled data, Med-MLLM exhibited competitive performance in these tasks.

\textbf{PeFoMed \cite{he2024pefomed}.} PeFoMed is a model designed for medical visual question answering (Med-VQA), employing a parameter-efficient fine-tuning approach aimed at enhancing the applicability of LLMs in multimodal environments. The model's design objective is to address the limitations of traditional classification tasks in handling open-ended questions by adopting a generative task to answer questions. PeFoMed's core idea is to use a pre-trained MLLM as the base model and adjust it using parameter-efficient fine-tuning techniques to meet the specific requirements of the Med-VQA task. During training, the model freezes the visual encoder and LLM, updating only the visual projection layer and low-rank adaptation layer, significantly reducing the number of parameters that need to be trained and lowering the computational resource demands. To enhance performance, PeFoMed employs a two-stage fine-tuning strategy with specific prompt templates. In the first stage, the model undergoes fine-tuning with large-scale multimodal data from a general domain to acquire basic capabilities for multimodal tasks. In the second stage, the model is fine-tuned with data consisting of medical images and text pairs in order to excel in the Med-VQA task. Manual evaluations were used to assess the model's performance and compare it to other models. The results demonstrate that PeFoMed achieved an overall accuracy of 81.9\% on closed-ended questions, a 26\% absolute accuracy improvement over the benchmark model GPT-4v. PeFoMed's contributions are notable in several aspects. It not only introduces a parameter-efficient fine-tuning method, but also enables LLMs to adapt to the Med-VQA task under limited resource and dataset conditions. Moreover, it designs specific prompt templates and a two-stage fine-tuning strategy to improve the model's performance and adaptability. Through experiments on public benchmark datasets, PeFoMed has proven to exhibit outstanding performance among generative Med-VQA models.

\section{Double-edged sword of LLM for Medcine} \label{Double-edged}

As we delve into the double-edged sword of LLMs for medicine, it becomes imperative to explore both the benefits they offer in reshaping healthcare paradigms and the deficiencies and challenges they confront in their integration into the healthcare ecosystem. Let us navigate through the multifaceted landscape of LLMs to discern their impact on the healthcare landscape. The benefits and challenges are shown in Table \ref{table:benefits&challenges}.

\begin{table}[ht]
    \centering
	\caption{Benefits and challenges medical LLM.}
	\label{table:benefits&challenges}
    \footnotesize
    \renewcommand{\arraystretch}{1.15}
	\begin{tabularx}{0.47\textwidth}{m{1.5cm}<{\centering}|m{5.9cm}<{\raggedright}}
		\toprule
		\hline
		\textbf{Domain}& \multicolumn{1}{c}{\textbf{Decriptions}} \\  
        \hline

        \multirow{13}{*}{Benefits} & LLM brings about more accurate diagnosis and prediction in medicine, promoting early disease detection and personalized treatment planning. \\
        \cline{2-2}
        
        & LLM integrates the latest medical advancements, providing real-time decision support and updated medical knowledge to physicians, thus optimizing clinical decision-making processes.  \\
        \cline{2-2}
        
        & LLM can personalize treatment plans and facilitate drug development, offering patients more precise treatment options and medication choices. \\

        \cline{2-2}
        
        & LLM enhances patient management and healthcare processes, improving medical efficiency and the quality of patient care. \\

        \cline{2-2}

        & LLM supports medical education and dissemination of healthcare knowledge, fostering continuous learning and improvement among medical students and practitioners.\\

        \hline

        \multirow{9}{*}{Challenges}
        & LLM faces fundamental challenges in medical applications due to high computational resource demands and low efficiency.  \\
        
        \cline{2-2}

        & Issues such as data imbalance, privacy protection, and data quality pose challenges to the applications of LLM in healthcare. \\
        
        \cline{2-2}

        & LLM encounters practical application challenges in clinical validation, multilingual adaptation, and multicultural adaptation. \\

        \cline{2-2}
        
        & The design of LLM needs to consider ethical concerns such as bias and fairness, privacy protection, transparency, and accountability. \\
        
        \hline

        \bottomrule
        \end{tabularx}
\end{table}

\subsection{Benefits of LLM for medicine}

LLMs hold significant promise for strengthening the healthcare landscape, offering advantages that can revolutionize medical practices. LLMs have multifaceted capabilities that span from augmenting diagnostic precision and enabling personalized treatment regimens to facilitating real-time access to cutting-edge medical knowledge. These advantages emphasize their pivotal role in reshaping healthcare paradigms and fostering enhanced patient-centric care. Here are several key benefits of LLM for medicine.

\subsubsection{Enhanced capabilities of diagnosis and prediction}

\textbf{Early detection and prediction.} LLMs play an essential role in the timely detection of disease indicators or factors predisposing individuals to health risks. For example, through analysis of extensive clinical records \cite{yang2022large} and various imaging modalities, LLMs possess the capability to prognosticate the likelihood of cardiac events or anticipate the probability of diabetic complications in patients. Their comprehensive assessment amalgamates multifaceted data points, allowing for nuanced risk stratification and early intervention strategies tailored to individual patient profiles.

\textbf{Prediction of treatment efficacy.} LLMs can anticipate and assess the efficacy of particular treatment protocols customized to individual patients. For instance, by harnessing the wealth of patient records and genomic information, LLMs can prognosticate survival rates or gauge the anticipated responses to specific cancer therapies \cite{iannantuono2023applications}. By amalgamating diverse datasets, LLMs enable precise treatment prognoses, facilitating informed decision-making tailored to the unique characteristics of each patient's medical profile. This capability offers invaluable insights into treatment outcomes, contributing to more targeted and effective therapeutic strategies in oncology and beyond.

\subsubsection{Knowledge integration and real-time information access}

\textbf{Latest medical advancements.} LLMs amalgamate a wealth of global medical literature, comprehensive clinical trial data, and expert insights, offering a dynamic platform for real-time updates in medical knowledge. For instance, LLMs can dynamically refine and revise cancer treatment protocols to align with the most recent clinical trial findings and evolving therapeutic guidelines \cite{yuan2023advanced}. By continuously synthesizing and analyzing an extensive array of sources, LLMs facilitate an agile and responsive framework for medical professionals, ensuring that treatment strategies stay abreast of the rapidly evolving landscape of evidence-based medicine. This adaptability helps to optimize patient care by incorporating the most current and relevant information into clinical decision-making processes.

\textbf{Clinical decision support.} LLMs serve as real-time decision support systems, significantly augmenting the depth and precision of diagnostic processes and treatment strategies for clinicians \cite{benary2023leveraging}. By swiftly processing vast volumes of patient data alongside the latest medical insights, LLMs offer immediate and comprehensive guidance to healthcare professionals. This real-time assistance bolsters the accuracy of diagnoses and treatment plans, empowering clinicians to make more informed and nuanced decisions at the point of care. The seamless integration of cutting-edge knowledge into clinical practice elevates the quality of healthcare delivery, ensuring that medical interventions align with the most current and validated information available in the field.

\subsubsection{Personalized treatment and drug development}

\textbf{Tailored treatment plans.} Leveraging individual patient data, LLMs can specialize in crafting meticulously tailored treatment blueprints encompassing an array of personalized interventions \cite{stade2023large}. These comprehensive plans cover everything from selecting the most appropriate medications to fine-tuning dosage parameters and even assisting in surgical strategies. For instance, drawing insights from detailed genomic profiles and comprehensive medical histories, LLMs excel at charting precise, individualized therapeutic routes for cancer patients. By scrutinizing intricate genetic information and patient-specific medical trajectories, LLMs contribute to the development of highly targeted therapy plans, optimizing treatment efficacy while minimizing potential adverse effects. This personalized approach stands as a testament to the potential of LLMs to revolutionize patient-centric care in oncology and beyond.

\textbf{Drug development and precision medicine.} LLMs excel at forecasting medication effectiveness and anticipating potential side effects, thereby expediting the drug development process and nurturing the realm of precision medicine. For instance, they undertake the pivotal task of predicting how a drug would perform in synergy \cite{li2024cancergpt}, significantly contributing to the meticulous design of more targeted and refined clinical trials. By delving into extensive datasets and intricate biological markers, LLMs offer invaluable insights into the anticipated efficacy of medications across diverse patient populations. This predictive capability not only accelerates the drug discovery phase but also aids in tailoring clinical trials to specific subsets of patients, fostering a more nuanced and individualized approach toward developing new therapeutic interventions.

\subsubsection{Patient management and healthcare process optimization}

\textbf{Personalized patient management.} LLMs specialize in crafting highly customized health management strategies by intricately analyzing biological markers and lifestyle data. For example, they meticulously design preventive and management protocols tailored specifically for patients dealing with cardiovascular diseases \cite{gala2023utility}. By integrating detailed genomic profiles with comprehensive lifestyle habits, LLMs formulate precise plans aimed at preventing the onset or progression of cardiovascular conditions \cite{arslan2023exploring}. Leveraging this amalgamation of genetic predispositions and individual behaviors, they develop personalized strategies encompassing dietary recommendations, exercise regimens, and medication plans. This tailored approach not only addresses immediate health concerns but also empowers patients with personalized insights to proactively manage their cardiovascular health and mitigate future risks.

\textbf{Optimization of healthcare processes.} LLMs are important parts of enhancing efficiency in medical processes for improved patient care. They work by meticulously identifying potential bottlenecks within diagnostic and treatment workflows. For instance, they are able to recognize critical points of congestion or inefficiency and assist in diagnostic assessments \cite{yang2023performance}. Following this identification, LLMs provide valuable optimization suggestions for streamlining and expediting patient care delivery. Their insights could range from refining scheduling protocols to suggesting workflow modifications that enhance the overall efficacy and promptness of healthcare services. By targeting inefficiencies and optimizing workflows, LLMs significantly contribute to the seamless and expedited delivery of quality healthcare, ultimately benefiting patient outcomes.

\subsubsection{Clinical education and dissemination of medical knowledge}

\textbf{Medical education and training.} The advanced LLMs serve as instrumental tools in the realm of medical education, providing a rich repository of the most recent medical knowledge and comprehensive case studies tailored for both aspiring medical students and seasoned practitioners. Their dynamic functionalities enable the simulation of real-world clinical scenarios, offering an immersive learning experience that significantly contributes to the enhancement of clinical decision-making skills \cite{safranek2023role}. By leveraging LLMs, medical students gain access to a vast array of intricate case studies and real-time medical data, replicating authentic patient encounters. This hands-on exposure allows them to sharpen their analytical skills, expand their medical knowledge base, and refine their diagnostic and treatment planning abilities within a safe, simulated environment. Moreover, for practicing healthcare professionals, LLMs serve as invaluable resources for continuous learning and staying updated with the latest advancements in medicine. They provide a platform for honing diagnostic acumen and exploring diverse treatment approaches through interactive scenarios. Ultimately, these educational tools foster a culture of ongoing learning and skill development. It ensures that, in an ever-evolving healthcare landscape, medical practitioners are equipped with the expertise to deliver optimal patient care.

\textbf{Dissemination of medical knowledge.} LLMs empower patients by providing direct, comprehensive disease prevention and management guidance through user-friendly healthcare applications. For instance, applications leverage the vast knowledge base of LLMs to give patients tailored advice on disease prevention strategies \cite{zeng2020meddialog}. They offer insights into lifestyle modifications, preventive measures, and early warning signs associated with various medical conditions. This guidance is personalized, considering individual health profiles and promoting proactive health management among users. Moreover, LLMs assist in delivering meticulous disease management recommendations to patients already diagnosed with specific conditions. They provide detailed information on treatment adherence, and such information is expected to improve \cite{jin2024guard}. By harnessing the extensive medical knowledge encapsulated within LLMs, these healthcare applications bridge the gap between medical expertise and patient understanding. They empower individuals to make informed decisions regarding their health, fostering a proactive approach toward disease prevention and self-care management.

\subsection{Deficiencies and challenges of LLM for medicine} \label{sec:challenges}

For healthcare, the potential of LLMs, powered by artificial intelligence, is undeniably transformative. These models can analyze vast volumes of medical data, aiding in diagnosis, treatment recommendations, and medical research. However, when integrating into the healthcare ecosystem, they also encounter hurdles. The use of LLMs in healthcare poses a multifaceted challenge that necessitates careful consideration and solutions, from computational demands to ethical considerations. These challenges can be categorized into four fundamental domains: fundamental requirements, data-related issues, practical applications, and ethical concerns. Thus, delving into these challenges can help us better understand the complexities of deploying LLMs in the healthcare field.

\subsubsection{Fundamental requirements}

\textbf{Computational resources.} LLMs in healthcare often lack the computational efficiency required for real-time processing. These resource-intensive models can result in longer response times, rendering them less suitable for applications that demand immediate decisions, such as emergency diagnosis and treatment. Furthermore, models like GPT-3 \cite{ye2023comprehensive}, GPT-4 \cite{achiam2023gpt}, or custom medical models require powerful computing hardware, typically in the form of graphics processing units (GPUs) or specialized hardware like tensor processing units (TPUs). With an enormous number of parameters, these models demand significant computational power for both training and inference. Acquiring and maintaining the necessary computational resources can incur prohibitively high costs. This encompasses not only the hardware itself but also the electricity and cooling systems essential for efficient operation. Healthcare institutions and research organizations must allocate budgets to cover these substantial expenses.

\textbf{Model efficiency.} In terms of data utilization, LLMs must effectively leverage the available information, especially in healthcare, where data can be scarce and sensitive. Inefficient models may necessitate more data than is practically obtainable, or they may fail to yield meaningful insights from limited datasets. Additionally, LLMs are susceptible to overfitting when training on limited data \cite{dodge2020fine}, which can undermine their ability to generalize to new and diverse medical cases. The process of optimizing inference is a critical component of LLMs. While model training demands substantial resources, the efficiency of inference holds equal importance. Healthcare applications require models to deliver timely responses, making the speed of inference a critical consideration. Efficient inference methods are essential to meet the demands of real-time decision-making in healthcare settings.

\subsubsection{Data-related issues} 

\textbf{Data privacy \cite{gui2024privacy}.} Healthcare data contains sensitive patient information, such as medical records, imaging data, and personal identifiers. Mishandling patient data can result in data breaches, identity theft, and legal repercussions. Ensuring that LLMs handle patient data with strict encryption, access controls, and audit trails is crucial to protecting patient confidentiality.

\textbf{Data quality.} Healthcare data comes in diverse formats, including structured electronic health records (EHRs), unstructured text reports, medical images, and time-series data. Ensuring data quality across these varied data types is challenging due to potential inconsistencies, errors, and data source variations. For example, in medical image analysis, it is critical to ensure the precision of annotations, such as identifying tumors in medical images \cite{osborne2009annotating}. Inaccurate annotations can lead to erroneous model training and subsequent misdiagnoses.

\textbf{Interpretability and trustworthiness.} In the healthcare domain, LLMs' interpretability and explainability are critical for their acceptance and utility among healthcare professionals. These models must provide clear explanations for decisions, especially in diagnoses and treatment recommendations. By understanding the reasoning behind these decisions, healthcare professionals can evaluate the model's outputs, identify biases or limitations, and make informed decisions based on the information provided. The transparency offered by interpretability enhances collaboration, allowing healthcare professionals to engage in meaningful discussions with the model, seek clarification, and ultimately improve healthcare decision-making. The interpretability and explainability of LLMs in healthcare not only build trust but also ensure their effective support in delivering high-quality care to patients.

\subsubsection{Practical applications}

\textbf{Clinical validation.} Conducting robust clinical trials to validate healthcare models is a challenging endeavor, primarily due to the inherent variability within medical data. Patient characteristics, medical conditions, and treatment approaches exhibit significant differences, underscoring the need to effectively address this diversity when validating the model's performance. Furthermore, scaling clinical validation to encompass a broad and diverse patient population while involving multiple healthcare institutions is a complex undertaking. It entails not only logistical challenges but also the critical task of ensuring that the model consistently performs effectively across diverse clinical settings and demographics.

\textbf{Multilingual and multicultural adaptation.} Adapting healthcare models to multiple languages necessitates a substantial investment in data collection, translation, and cross-lingual model training. This is due to the inherent variations in language structure and medical terminology across different languages, which can pose significant challenges. Furthermore, integrating cultural nuances into the model's understanding and recommendations can be a complex and nuanced process. This is because healthcare practices, beliefs, and patient expectations can vary significantly from one culture to another, requiring precise model adaptations for different cultural contexts. Adding to the complexity, healthcare practices and medical standards may differ across various regions. Consequently, adapting models to address these variations while ensuring consistent performance across diverse clinical settings presents a formidable undertaking.

\subsubsection{Ethical concerns}

\textbf{Bias and fairness.} Medical language models should be designed to mitigate bias and ensure fairness in medical decision-making. Biased models can lead to disparities in outcomes, disproportionately affecting underserved or minority populations. Ethical considerations involve conducting bias audits, adjusting models to reduce bias, and transparently addressing bias-related concerns.

\textbf{Privacy concerns.} The ethical use of medical models necessitates robust privacy protections. Ensuring that patient data is anonymized, encrypted, and stored securely is vital. Models should also be designed to minimize data exposure and limit access to sensitive information, all while maintaining the quality of medical insights.

\textbf{Accountability and transparency.} Ethical considerations include mechanisms for accountability and transparency in AI-driven medicine. This involves clear documentation of model decision-making, traceability, and the ability to explain the rationale behind specific medical recommendations. Transparent and interpretable models foster trust among healthcare professionals and patients.

\textbf{Patient autonomy.} Respecting patient autonomy is essential. Patients should have the right to accept or reject AI-driven recommendations, and healthcare providers should ensure that these recommendations align with patients' values, preferences, and informed consent.

\textbf{Preventing data misuse.} Ethical guidelines entail safeguarding medical models from misuse or exploitation. Preventing unintended consequences, such as inappropriate use of patient data, model hacking, or malicious intent, is a critical ethical consideration.

\section{Opportunities and Future Directions} \label{sec:oppotrunities}

\begin{figure*}[ht]
    \centering
    \includegraphics[clip,scale=0.3]{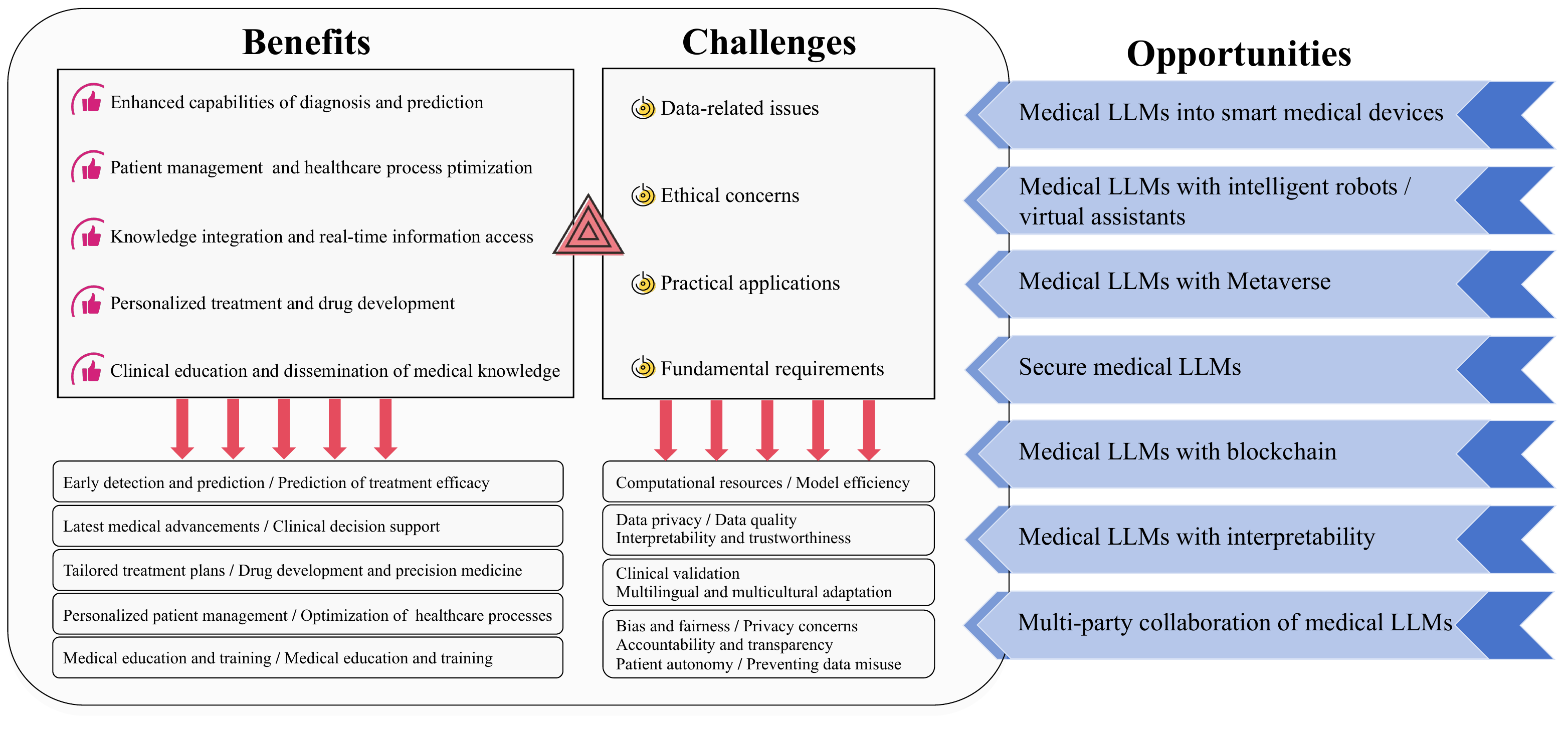}
    \caption{Benefits, challenges, and opportunities of medical LLMs.}
    \label{fig:cof}
\end{figure*}

Medical LLMs present extensive opportunities and future directions that will further drive innovation and improvement in medical practices. As shown in Figure \ref{fig:cof}, we give some potential opportunities and future directions to develop them, including medical LLMs into smart medical devices \cite{wang2019bio}, medical LLMs with intelligent robots/virtual assistants \cite{huang2023intelligent, preum2021review}, medical LLMs in Metaverse \cite{chen2022metaverse, chen2024metaverse}, secure medical LLMs \cite{he2023controlling}, medical LLMs with blockchain \cite{roman2018blockchain}, and multi-party collaboration of medical LLMs \cite{chen2023federated}.

\subsection{Medical LLMs into smart medical devices}

For real-time health monitoring and data interpretation \cite{li2023integrated}, the integration of LLMs into smart medical devices enables continuous monitoring of patient's physiological parameters. These medical LLMs can promptly interpret transmitted data, providing critical analysis and feedback, thereby facilitating early detection of potential health issues. This integration also should support remote medical consultation and monitoring, allowing healthcare professionals such as physicians and nurses to communicate remotely with patients via voice or text \cite{wu2023internet}. Effective remote monitoring is enabled by leveraging medical LLMs to offer medical advice and personalized recommendations. Intelligent medication management, facilitated by IoT technology, is another potential application. By integrating IoT devices with medical LLMs, medication packaging or smart medicine boxes can remind patients of medication schedules, provide explanations of medication information, and adjust medication plans based on patient feedback \cite{dou2024shennongmgs}, thereby enhancing medication adherence. In clinical settings, it is feasible to use IoT devices in operating rooms and emergency environments. Medical LLMs can interpret real-time surgical data and monitor device feedback, offering immediate intraoperative guidance or emergency advice \cite{raheja2023iot}. This can lead to improvements in surgical efficiency and the speed of emergency decision-making. Furthermore, in inpatient recovery monitoring and recommendations, leveraging IoT technology to monitor the usage of rehabilitation devices and transmit rehabilitation data to medical LLMs is advantageous. The language model can offer personalized rehabilitation suggestions and monitor recovery progress, enabling healthcare teams, including physicians, therapists, and caregivers, to remotely adjust treatment plans \cite{neo2024use}. Integrating these approaches improves the intelligence, real-time capabilities, and personalization of healthcare services and broadens the prospects for the healthcare field.

\subsection{Medical LLMs with intelligent robots/virtual assistants}

In the field of healthcare, intelligent robots and virtual assistants have emerged as essential technological tools for delivering personalized and efficient medical services \cite{huang2023intelligent}. The integration of medical LLMs enhances these intelligent entities' semantic understanding and NLP capabilities \cite{revell2024generative}. Various types of intelligent robots, such as surgical robots \cite{chen2024llm} and caregiving robots \cite{padmanabha2024voicepilot}, and virtual assistants \cite{dong2023towards}, including chatbots and voice assistants, can better comprehend and respond to patient needs by incorporating LLMs, offering more intelligent medical consultation, appointment scheduling, and health management services \cite{vu2024gptvoicetasker}. For instance, intelligent robots equipped with deep learning algorithms and medical LLMs can perform emotion recognition and semantic analysis, enabling a more precise understanding of patient language and emotions \cite{dong2023towards}. This enables them to provide personalized medical advice and services as needed. Similarly, virtual assistants can leverage the capabilities of medical LLMs to automate medical record-keeping and offer clinical decision support for healthcare professionals. This enhances work efficiency and diagnostic accuracy in clinical settings. Thus, the integration of medical LLMs with intelligent robots and virtual assistants promises to bring about more intelligent and efficient medical services, ultimately enhancing the medical experience and treatment outcomes.

\subsection{Medical LLMs with Metaverse}

In the Metaverse \cite{sun2022metaverse}, virtual medical assistants create an interactive healthcare experience for both doctors and patients. These assistants, powered by medical LLMs, can interact with patients in virtual reality or augmented reality, providing health information, conducting self-diagnoses, and explaining medical concepts. Additionally, leveraging the ability of Metaverse to construct virtual worlds, medical virtual spaces can be established for remote medical consultations and treatments. Doctors and patients can engage in face-to-face medical consultations within this virtual environment, with medical LLMs playing a role in explaining medical conditions and providing treatment recommendations \cite{chen2024open}. For medical students and professionals, Metaverse, especially human-centric Metaverse \cite{yang2023human}, offers opportunities for smart education \cite{lin2022metaverse}, virtual practice, enhancing medical training and simulation \cite{el2023integration}. Medical LLMs can serve as virtual mentors, offering real-time feedback, explaining surgical procedures, and addressing relevant questions, thus facilitating the training of healthcare professionals. In the long run, medical research and collaboration will also witness a new era within the Metaverse \cite{pressman2024ai}. Future research can focus on collaborating within shared virtual environments, and medical language models can assist in analyzing complex medical literature, providing research recommendations, and advancing medical research. The innovative integration of medical LLMs with Metaverse holds promise for the healthcare sector's future. By embedding medical LLMs into the Metaverse framework, we can establish more comprehensive, intelligent, and personalized healthcare services.

\subsection{Secure medical LLMs}

The use of medical LLMs in the healthcare industry necessitates ensuring data security and privacy. To safeguard the security of medical LLMs, multiple measures must be taken. Firstly, data collection and storage for medical LLMs must adhere to strict privacy protection standards, such as HIPAA \cite{marks2023ai} and GDPR \cite{lawlor2023impact}, to ensure the confidentiality of patients' personal information. Secondly, during the training and application process of medical LLMs, secure and controllable technical measures must be adopted to prevent data from being subjected to malicious attacks or misuse. For example, encryption algorithms and access control policies are used to ensure the security of data transmission and access. Additionally, comprehensive permission management and auditing mechanisms must be established for medical LLM applications to ensure that only authorized personnel can access and use relevant data and models. Finally, the applications of medical LLMs must comply with relevant laws, regulations, and ethical guidelines to protect patients' legal rights and medical information security. By comprehensively implementing these measures, medical LLMs can be effectively secured, providing reliable technical support for their applications in the healthcare field.

\subsection{Medical LLMs with blockchain}

Beginning with the utilization of blockchain's smart contract functionality, patients gain control over their medical data, assisted by medical LLMs as interpreters. This empowers selective data sharing and tracking usage. Furthermore, smart contracts manage processes like billing and insurance claims, with LLMs explaining contract content. Storing medical data on the blockchain enhances security and privacy. Medical LLMs ensure secure access and accurate analysis of patient records, contributing to transparency in clinical trial data. Moreover, constructing a medical knowledge graph on blockchain supports decision-making for doctors and researchers, with medical LLMs providing deeper insights and enhancing research credibility. Additionally, recording pharmaceutical and medical device information on the blockchain ensures authenticity and traceability. Medical LLMs play a crucial role in identifying counterfeit products, thus safeguarding patient safety \cite{heston2024prespective}.

\subsection{Medical LLMs with interpretability}

The interpretability of medical LLMs is a crucial topic in the current field of medical artificial intelligence research. In the process of enhancing model transparency and comprehensibility, a focus on both global and local interpretability is significant. In the realm of interpretable deep learning models, innovative methods such as attention mechanisms \cite{chen2020recurrent} and interpretable neural networks \cite{singh2023augmenting} are continuously emerging to provide more detailed explanations. Furthermore, interpretability directly influences the model's trustworthiness and reliability, prompting the need for research on how transparency and interpretability can enhance the model's reliability and reduce potential medical errors. While providing interpretability, privacy protection for patients must also be considered, involving the adopted privacy protection techniques such as differential privacy \cite{song2024llm}. Incorporating the knowledge of domain experts can be utilized to interpret model outputs, thereby increasing the credibility of the model's output. In the context of rule extraction, interpretability is relatively straightforward for decision tree models, allowing for the extraction of rules and information on nodes and branches. Furthermore, addressing uncertainties in model quantification, establishing interpretability standards and evaluation methods, and educating healthcare professionals on understanding and interpreting model outputs are all critical factors for ensuring the success of medical LLMs in terms of interpretability. Considering these aspects comprehensively can make medical LLMs more understandable and acceptable, thereby enhancing their usability and credibility in practical medical applications.

\subsection{Multi-party collaboration of medical LLMs}

The incorporation of medical LLMs within the healthcare domain necessitates concerted efforts from various stakeholders, including governmental bodies, healthcare institutions, patients, and research establishments. Governments wield a pivotal role in shaping policies, regulations, and standards governing the development and implementation of medical LLMs. They possess the capacity to mobilize resources, facilitate data exchange, and provide essential infrastructure support, such as computing capabilities, indispensable for both training and operationalizing these models. Healthcare institutions serve as primary arenas for the deployment of medical LLMs, given their status as providers of medical services. Collaborating with governmental bodies, they can establish platforms conducive to the sharing and integration of medical data, pivotal for refining and optimizing these models. Furthermore, governments and healthcare institutions can not only foster patient involvement in data collection and sharing by instituting mechanisms that promote active engagement, but also harness intelligent health management platforms to deliver personalized healthcare services, thereby improving patient experiences and treatment outcomes. Governments can bolster research establishments through financial support and collaborative frameworks, thereby catalyzing technological innovation and advancement in the realm of medical LLMs. Moreover, partnerships between healthcare and research establishments can drive exploration into the application of medical LLMs across disease diagnosis, prevention, and treatment, fostering innovation in medical technologies. Patients, as recipients of medical care, occupy a pivotal role in the utilization of medical LLMs. Research institutions, meanwhile, play a crucial role in the research and development of these models, often collaborating with healthcare establishments to explore their myriad applications and expedite their translation into practical medical solutions.

By fostering collaboration among governmental bodies, healthcare institutions, patients, and research establishments, the comprehensive integration of medical LLMs into the healthcare landscape can be achieved. This collaborative approach promises to deliver intelligent, personalized, and efficient healthcare solutions, thereby elevating healthcare standards and enhancing patient quality of life.

\section{Conclusion}  \label{sec:conclusion}

In this paper, we comprehensively explore the pivotal role of LLMs in the field of medicine. These models demonstrate significant potential not only in medicine-assisted diagnosis, biopharmaceutical design, and medical image segmentation, but also in achievements related to health management, doctor-patient communication, and multimodal applications of LLMs in medicine. However, challenges persist, encompassing issues such as data privacy, model interpretability, ethical concerns, and technical difficulties in practical implementations. Future research should focus on addressing these challenges to ensure the reliability and safety of models in real clinical environments. Regarding these problems, we proposed several technologies that are possible to combine with medical LLMs to solve them, including Metaverse, blockchain, smart medical devices, and future research directions for researchers. In summary, LLMs bring unprecedented opportunities to the field of medicine. LLMs in medicine are poised to play a greater role in personalized medicine, new drug development, and health management. Nevertheless, it is imperative to prioritize ethical and privacy considerations in this process. We anticipate achieving more significant accomplishments in improving patient quality of life, advancing medical research, and optimizing medical processes. Encouraging collaborative efforts among future researchers and practitioners is essential to drive the development of LLMs in medicine for the benefit of humanity.

~\\

\textbf{Acknowledgments} This research was supported in part by the National Natural Science Foundation of China (No. 62272196), the Natural Science Foundation of Guangdong Province (No. 2022A1515011861), Guangzhou Basic and Applied Basic Research Foundation (No. 2024A04J9971). 

\textbf{Author contributions} Yanxin Zheng: paper reading and review, writing original draft. Wensheng Gan: conceptualization, review and editing, supervisor. Zefeng Chen and Zhenlian Qi: conceptualization, review and editing. Qian Liang and Philip S. Yu: review and editing.

\textbf{Data availability} This is a review paper, and no data was generated during the study.

\textbf{Conflict of interest} The authors declare that they have no known competing financial interests or personal relationships that could have appeared to influence the work reported in this paper.

\bibliographystyle{elsarticle-num-names}

\bibliography{llm4medicine.bib}


\end{document}